\pdfoutput=1

\documentclass[11pt]{article}

\usepackage{ACL2023} 

\usepackage{times}
\usepackage{latexsym}

\usepackage[T1]{fontenc}

\usepackage[utf8]{inputenc}

\usepackage{microtype}

\usepackage{inconsolata}

\usepackage{algorithmic}
\usepackage{graphicx}

\usepackage{multirow}
\usepackage{makecell}
\usepackage{dirtytalk}
\usepackage{booktabs}

\usepackage{enumitem}
\usepackage{amsmath}
 
\usepackage{amssymb}
\usepackage[normalem]{ulem} 
\usepackage{multirow} 
\usepackage{subcaption} 
\usepackage{graphicx} 
\usepackage{arydshln} 

\usepackage{color}

\title{Towards Generalizable Generic Harmful Speech Datasets for Implicit Hate Speech Detection}

\author{\textbf{Saad Almohaimeed}, \textbf{Saleh Almohaimeed}, \textbf{Damla Turgut} and \textbf{Ladislau Bölöni} \\
Dept. of Computer Science, University of Central Florida \\
}

\begin{document}
\maketitle
\begin{abstract}
Implicit hate speech has recently emerged as a critical challenge for social media platforms. While much of the research has traditionally focused on harmful speech in general, the need for generalizable techniques to detect veiled and subtle forms of hate has become increasingly pressing. Based on lexicon analysis, we hypothesize that implicit hate speech is already present in publicly available harmful speech datasets but may not have been explicitly recognized or labeled by annotators. Additionally, crowdsourced datasets are prone to mislabeling due to the complexity of the task and often influenced by annotators' subjective interpretations. In this paper, we propose an approach to address the detection of implicit hate speech and enhance generalizability across diverse datasets by leveraging existing harmful speech datasets. Our method comprises three key components: influential sample identification, reannotation, and augmentation using Llama-3 70B and GPT-4o. Experimental results demonstrate the effectiveness of our approach in improving implicit hate detection, achieving a +12.9-point F1 score improvement compared to the baseline.
\end{abstract}

\section{Introduction}

The field of harmful speech classification has garnered significant attention, with extensive research addressing various aspects of this phenomenon. Several studies focus on general hate speech, such as~\cite{davidson2017automated},~\cite{OLID-zampieri-etal-2019-predicting},~\cite{hatexplain-mathew2021}. Others have delved into specific forms of hate speech, including works by~\cite{waseem-hovy:2016:N16-2},~\cite{founta2018large}, and~\cite{ousidhoum-etal-2019-multilingual}. For clarity, these datasets will be referred to as generic datasets throughout this paper.

While many of these datasets include annotated examples of implicit hate speech, the reliance on crowdsourced annotators has introduced variability in labeling, with some annotators identifying implicit hate as harmful, while others do not. In contrast, publicly available datasets explicitly focused on implicit hate speech (specialized datasets) are far fewer than their generic counterparts. Leveraging the extensive data available in generic datasets and reformatting them to enhance generalizability in implicit hate detection presents a promising opportunity.

To explore this, we analyzed four generic datasets—Davidson~\cite{davidson2017automated}, HateXplain~\cite{hatexplain-mathew2021}, Waseem~\cite{waseem-hovy:2016:N16-2}, and Founta~\cite{founta2018large}—using an offensive language lexicon developed by~\cite{almohaimeed2024transfer}. This lexicon, comprising 1.8k offensive terms, was employed to calculate the proportion of positive samples (annotated as harmful) that were free of offensive language. Such samples were presumed to indicate implicit hate. The results revealed that the percentage of positive samples free of offensive language was 71.7\%, 14\%, 36.2\%, and 20.8\% for Waseem, Davidson, Founta, and HateXplain, respectively. This analysis is contingent on the lexicon's comprehensiveness; some rare offensive terms may not be included, potentially classifying certain offensive samples as implicit hate. Given these findings, we propose an approach to guide models trained on generic datasets toward better generalizability for implicit hate detection while preserving their ability to identify explicit harmful speech. 

Harmful speech, as defined in this paper, encompasses any expression that includes explicit hate, implicit hate, offensive language, or any content that may cause harm to the reader or the target. Furthermore, implicit hate is a subtype of harmful speech, characterized by hate conveyed in a veiled or subtle manner.

The key contributions of this paper are as follows:
\begin{itemize} \item Introduce a novel approach to generalize high-level (general-purpose) datasets to specialized classes that exist within these datasets but lack explicit annotations.
\item Develop a trusted samples dataset comprising 500 samples, designed to serve as a benchmark for evaluating various types of harmful speech.
\item Apply the proposed generalization approach to adapt harmful speech datasets for the task of implicit hate classification.
\item Demonstrate the utility of influential sample identification by training classifiers using three proposed configurations across four different hate speech datasets and evaluating their performance on seven datasets, using cross-dataset settings.
\end{itemize}

\section{Related Work}

\subsection{Generalizable Implicit Hate Classifier}

There are several research studies that aimed to generalize implicit hate datasets in cross-dataset evaluation settings. Some of them proposed techniques to push positive samples towards their corresponding implications along with augmentation techniques~\cite{kim-etal-2022-generalizable}, bring the encoding of the explicit and implicit hate samples that share the same target close to each other~\cite{ocampo-etal-2023-unmasking}, proposed a pre-trained model on ToxiGen~\cite{hartvigsen-etal-2022-toxigen}, integrated with prompting techniques~\cite{kim-etal-2023-conprompt}, push the embeddings of positive samples to the closest positive cluster~\cite{almohaimeed2025CPC} or push samples that share the same semantics into the same embedding cluster~\cite{ahn-etal-2024-sharedcon}. All these studies used a limited number of implicit hate datasets, such as ToxiGen~\cite{hartvigsen-etal-2022-toxigen}, IHC~\cite{IHC-elsherief-etal-2021-latent}, DYNA~\cite{DYNA-vidgen-etal-2021-learning}, SBIC~\cite{sap-etal-2020-social} for training implicit hate detection models.

\subsection{Identification of Influential and Noisy Samples}

Most approaches for the identification of influential samples in machine learning rely on the loss function~\cite{Influ-koh2017, pruthi2020estimating, labelcorrection-jinadu2024noise}. These methods typically aim to mitigate the impact of influential samples during training, by either reweighting their loss or by removing them from the training dataset.

\cite{labelcorrection-jinadu2024noise} proposed a method for detecting and correcting mislabeled samples in machine learning. Their approach relies on loss correction within a multi-task learning framework, including a case study focusing on a hate speech dataset. The core idea behind multi-task learning in their methodology is to separate the predictions for each label based on the perspectives of individual annotators. The authors demonstrated the effectiveness of their approach, reporting an approximate 10-point improvement in the F1 score. This improvement was observed both with and without noise injection, where baseline performance dropped significantly under noise, thereby highlighting the robustness of their method.

\cite{noisy-early-learning-regularization-Liu-2020} observed that machine learning models tend to learn from correctly labeled data during early epochs but gradually begin to incorporate noisy labels in later epochs. To address this issue, they proposed a method that combines a regularization technique with a semi-supervised model. Their approach estimates the probability of the target label, enabling better generalization by avoiding overfitting and the memorization of noisy labels. Unlike early-stopping techniques that simply monitor the model during initial stages and halt training prematurely, their method emphasizes sustained learning without overfitting. Experimental results on CIFAR-10 and CIFAR-100 demonstrated that their approach achieves performance comparable to SOTA methods.

\cite{pruthi2020estimating} introduced TracIn, a method for calculating the influence of each training sample based on the model's predictions for test data. This approach operates in a multi-model setting as it relies on the loss function and gradients. Their experiments spanned image and text classification tasks, as well as one regression task. The authors observed that mislabeled text samples consistently exhibited high loss values, and TracIn effectively identified these mislabeled samples in the early epochs, outperforming other SOTA methods in speed and efficiency. Their method demonstrated practical benefits, improving the accuracy on the CIFAR-10 dataset by 2 percentage points and the MNIST dataset by approximately 1 percentage point.

\cite{noisy-arazo2019unsupervised} introduced dynamic bootstrapping, an enhancement of the static bootstrapping technique developed by~\cite{noisy-reed2015training}, to address the challenge of noisy labels in image datasets during training. Unlike the static approach, dynamic bootstrapping adapts individually to each sample, providing a more flexible mechanism to avoid overfitting noisy labels. The authors also combined their method with a modified version of the data augmentation technique proposed by~\cite{zhang2018mixup}, adapting it to operate dynamically for each sample. Experiments conducted on CIFAR-10, CIFAR-100, and TinyImageNet showed that dynamic bootstrapping effectively mitigated the impact of noisy labels and outperformed existing methods in generalization, leading to improved performance across these datasets.

\cite{Influ-han-tsvetkov-2020-fortifying} proposed a method to enhance implicit hate classifiers without requiring large annotated datasets. Their approach utilized probing examples from the SBIC dataset~\cite{sap-etal-2020-social} and employed several tracking methods to identify influential samples that contributed to misclassifications. These influential samples were then reannotated to improve the quality of the training data. The study demonstrated that identifying and reannotating mislabeled training samples significantly enhanced the model's performance. Among the methods tested, the gradient product proved to be the most effective technique for detecting influential samples.

We observed that most studies in the field of influential samples and noisy labels are applied on image classification tasks. However, several studies consider text classification such as~\cite{labelcorrection-jinadu2024noise} and \cite{Influ-han-tsvetkov-2020-fortifying}.

\section{Methodology}

\begin{figure}  
    \centering
    \includegraphics[width=0.5\textwidth]{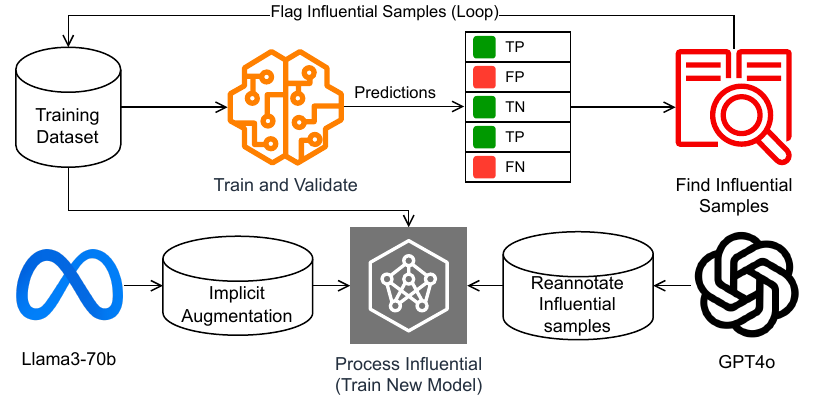} 
    \caption{The pipeline of our proposed methodology}
    \label{fig:pipeline_fig}
\end{figure}

\subsection{Datasets}

The scarcity of implicit hate datasets motivated us to leverage existing generic harmful speech datasets to improve generalizability for implicit hate detection. To address this, we utilized the generic datasets for training. These datasets serve as the foundation for applying our approach to enhance the detection of implicit hate.

We evaluated the trained model on three specialized implicit hate datasets: IHC~\cite{IHC-elsherief-etal-2021-latent}, OLID\_IH~\cite{OLID_IH-caselli-etal-2020-feel}, and THOS\_IH~\cite{THOS-almohaimeed}. Additionally, we cross-tested on the generic datasets to ensure that the performance on explicit harmful speech detection was not compromised by our approach. 

As a note, the Founta dataset as used here is shorter than its original published size from Founta et al.~\cite{founta2018large} due to data unavailability while fetching them using their published IDs through the Twitter's (X) API. Approximately 35k samples were removed as a result of platform policies or author decisions. Additionally, the labels in all datasets were unified into binary classes (normal or harmful) for generic datasets and (not implicit hate or implicit hate) for specialized datasets following the standardization introduced in~\cite{almohaimeed2024transfer}.

\subsection{Trusted Samples Dataset}

Our objective is to create a dataset that is significantly smaller than typical datasets but annotated with a level of effort and consideration that would not be feasible for a larger dataset. To achieve this, we define the Trusted Samples Dataset (TSD) as a benchmark testing dataset. Our intention is for this dataset to serve as a guide for improving data annotation to enhance generalizability and identifying influential data within the training dataset.

The TSD was constructed collaboratively using GPT-4o and two human experts in the field of harmful speech detection. It consists of 500 samples, evenly split between positive (harmful) and negative (non-harmful) samples. The positive samples were curated to encompass various types of harmful speech, including implicit hate, explicit hate, offensive language, racism, and sexism. These samples target both named and unnamed entities, covering a wide range of vulnerable and non-vulnerable groups as well as individuals. The GPT-4o generation for the required content was conducted between October 8th and 12th, 2024.

The inclusion of GPT-4o in the TSD creation process was critical for achieving a comprehensive and balanced dataset. GPT-4o contributed by identifying diverse targets that human experts might overlook and generating neutral samples that closely resemble harmful speech in structure and tone. This approach provides a robust metric for evaluation, as the TSD includes challenging examples that prevent the model from favoring a specific class (positive or negative) based solely on target presence.

\subsection{Influential Sample Identification} \label{sec:Influential_Identification}

Influential samples can be classified into two categories. The first category consists of samples that were mislabeled by the annotators, leading to discrepancies in model performance. The second category includes samples that closely resemble misclassified TSD examples but belong to the opposite class. Our objective is twofold: to correct mislabeled samples from the first category and to augment the second category to improve the model's generalizability. As depicted in Fig.~\ref{fig:pipeline_fig}, influential sample identification involves finding the samples responsible for the misclassification of each TSD sample.

Let us assume $D = \{t_1, t_2, t_3, \dots, t_n\}$ where $D$ is the training dataset and $t_i$ represents a text sample in $D$. Additionally, assume $TSD = \{gt_1, gt_2, gt_3, \dots, g_m\}$ where $TSD$ is the trusted samples dataset and $gt_j$ represents a text sample in $TSD$.

Let $M$ be a trained model using $D$, and let $\hat{y_j}$ denote $M$'s prediction for $gt_j$, defined $M(\text{gt}_j) = \hat{y_j}, \quad \text{for } j \in \{1, 2, \dots, m\}$. We define the set of misclassified $TSD$ samples by $M$ as $E$ such that $E = \{\text{gt}_j \in \text{TSD} \mid \hat{y}_j \neq y_j\}$.

Next, let us define a cosine similarity metric as 
$csim(t_i, gt_j) =\frac{e_{t_i} e_{gt_i}}{\|e_{t_i} \| \cdot \|e_{gt_i}\|}$, where $e$ represents the embedding of a given text sample produced by $M$. Finally, we define the top $x$ influential samples for $\text{gt}_j \in \text{E}$ as follows: 
\begin{align*}
&\text{top\_influence}(E, x) = \\ &\bigcup_{gt_j \in E} \text{Top-}x\big\{ csim(t_i, gt_j)  \mid t_i \in D,  y_i = \hat{y_j} \big\}
\end{align*}

\subsection{GPT4 Annotation} \label{sec:gpt4_annotation_in_methodology}

Sometimes what makes a sample influential in making $M$ misclassify a given $gt_j$ is the fact that the sample was mislabeled by the annotators. It is difficult to engage humans in the loop of our pipeline Fig.~\ref{fig:pipeline_fig} to reannotate influential samples since we are dealing with an extensive amount of data, 107.8k rows in total from the 4 generic datasets along with full training process on each loop in our pipeline. So, it is necessary to find an applicable technique to reannotate influential samples such that it will be fixed in case it was mislabeled. Despite the limitations of GPT4 to overlook the explicit and implicit hate that targets individuals or to be confused with sarcasms and opinions~\cite{almohaimeed2024transfer}, several studies~\cite{almohaimeed2024transfer, gpt-Huang2023IsChatGPT,gpt4-donmez2024please} explain the effectiveness of GPT family on identifying hate text. \cite{gpt4-donmez2024please} studies the effectiveness and limitations of 16 LLMs on identifying the harmful content, and despite the limitations, the authors found that the GPT family (GPT3.5-turbo and GPT4) were the best closest performing LLMs to the human baseline. Taking these results into consideration, we employed GPT4o to reannotate the influential samples, and adjust the ground truth label if it does not match the GPT4o annotation. 

\subsection{Llama-3 Augmentation} \label{sec:Llama_augmentation_in_methodology}

We utilized the capabilities of open-source large language models to augment generic datasets with more implicit hate speech samples. Specifically, we employed Llama-3 8B and 70B~\cite{Llama-dubey2024llama} to paraphrase explicit harmful speech samples into implicit hate speech.

Llama-3 8B struggled to perform the task effectively, often repeating the provided system and user prompts instead of generating paraphrased outputs. In contrast, Llama-3 70B demonstrated a substantial understanding of the task and consistently produced accurate paraphrases. For this augmentation process, we used the 4-bit quantization setting on an NVIDIA H100 80GB GPU. The paraphrasing of a single harmful speech sample took an average of 9.58 seconds, resulting in a total runtime of approximately 136 hours to augment 51k harmful speech samples.

In the experiments, we explored several techniques to incorporate the augmented data into model training. The first approach duplicated each row with its augmented version, assigning the same label (positive). The second approach replaced explicit harmful speech rows with their implicit counterparts. However, in both cases, the BERT model struggled to learn from the augmented data and failed to converge, even when varying model hyperparameters.

Our observations indicate that models often face difficulties when training on datasets with a high proportion of deeply implicit samples — samples that even humans may find challenging to identify as harmful. Examples include statements like \textit{“RACE should stay in their own neighborhoods”} or \textit{“COUNTRY people need to take more responsibility”}. While such statements may not appear harmful to some, they can be deeply offensive to individuals belonging to vulnerable groups. These challenges are compounded for small models (e.g. BERT), which may lack the capacity to discern the implicit harmfulness of such texts.

In our case, we limited augmentation to the identified influential samples—a smaller subset of the dataset—which yielded significant performance improvements.

\section{Experimental Settings}\label{sec:Experiment_Settings}

For our experiments settings, we utilized BERT\textsubscript{base} uncased, which comprises 110 million parameters and 12 encoder layers~\cite{devlin2019bert}. We experimented with batch sizes of 4, 8, and 16 and learning rates ranging from 1e-6 to 1e-7. Although all experiments were run for 20 epochs, the reported results are based on the best epoch performance. 

\subsection{Experimental Setup}

\textbf{Text Preprocessing:} We applied several steps to clean and standardize the input data for the training model. These steps included removing URLs, user references (e.g., @USER), and hashtags from the text. We also decomposed contractions (e.g., doesn't~\textrightarrow~does not) and eliminated extra whitespaces. These preprocessing steps were applied to ensure consistency and reduce noise in the dataset.

\medskip

\noindent\textbf{Generic Datasets:} This category comprised four experiments using generic datasets for training. In these experiments, we focused on influential sample identification and processing. Each dataset went through multiple full training loops. During each loop, influential samples were identified, removed from the dataset, and then the model was retrained. The number of loops required to achieve the best results varied across datasets. Table~\ref{tab:generic_datasets_looping_results} shows the number of loops and the number of dropped samples for each dataset. The number of dropped samples represents the total count of the top $x$ samples for each misclassified $gt_i$. It is important to note that that the top $x$ samples for $gt_a$ and $gt_b$ may sometimes overlap. For Founta, we chose to drop the top 20 samples for each misclassified $gt_j$ sample per loop instead of 10, given its larger dataset size relative to the other generic datasets.

\medskip

\begin{table}[h!]
\caption{Influential Sample Identification Results of the Generic Datasets }\label{tab:generic_datasets_looping_results}
\centering
\resizebox{\columnwidth}{!}{%
\begin{tabular}{lllll}
\toprule
\textbf{Dataset} & \textbf{\# Top} & \textbf{\# Loops} & \shortstack{\textbf{Original} \\ \textbf{Size}}  & \shortstack{\textbf{\# Influential} \\ \textbf{Samples}} \\
\midrule
Waseem        & 10      & 16  & 16,907 & 8,343 \\ 
Davidson      & 10      & 3   & 24,783 & 1,434 \\ 
Founta        & 20      & 13  & 45,982 & 12,770 \\ 
HateXplain    & 10      & 7   & 20,148 & 2,328 \\ 
\bottomrule
\end{tabular}
}
\end{table}

\noindent \textbf{Specialized Datasets:} This category involved three experiments where the model was trained on specialized datasets (i.e. implicit hate datasets) and tested across other specialized datasets. The objective was to compare the effectiveness of generalizing from generic datasets toward implicit hate detection versus using datasets explicitly specialized for implicit hate.

\medskip

\noindent\textbf{Testing:} For each test across the seven datasets, the results presented in Tables~\ref{tab:train_with_harmful_test_with_harmful},~\ref{tab:train_with_harmful_test_with_implicit} and~\ref{tab:train_with_implicit_test_with_implicit} represent the average performance across five run with different random seeds. Each run used balanced random samples consisting of 500 positive and 500 negative examples from the test set. This approach ensured consistency and robustness for evaluating the results across all experiments.

\subsection{Metrics}

The metrics used in all experiments were the F1 score and Recall. While the F1 score is a more comprehensive metric as it incorporates Recall in its calculation, Recall was also included as an additional metric to ensure fairness when comparing the performance of generic datasets (Table~\ref{tab:train_with_harmful_test_with_implicit}) against specialized datasets (Table~\ref{tab:train_with_implicit_test_with_implicit}). This consideration stems from the inherent differences in the annotation approaches: generic datasets label all harmful speech samples as positive, whereas specialized datasets label only implicit hate samples as positive. Consequently, other types of harmful speech (e.g., explicit hate and offensive language) are annotated as negative in specialized datasets, alongside neutral speech. As a result, when a model trained on generic datasets is tested on specialized datasets, it is likely to predict many explicit harmful speech samples as positive, resulting in a higher number of False Positives (FP). This discrepancy occurs because specialized datasets annotate such samples as negative, focusing solely on implicit hate. Given this context, Recall is particularly suitable as it measures the True Positive (TP) rate, capturing the proportion of correct positive predictions relative to the total number of actual positive samples. In our case, Recall provides a reliable measure of the model's ability to identify implicit hate samples within the specialized datasets, providing a fair and meaningful comparison. However, we also include the F1 score in Tables~\ref{tab:train_with_harmful_test_with_implicit} and~\ref{tab:train_with_implicit_test_with_implicit} to ensure a balanced view of the model's performance without overemphasizing Recall at the expense of Precision.

\section{Experiments and Results}\label{sec:Experiments&Results}

\begin{table*}[!ht]
\caption{Performance of generic datasets evaluated across each other. Baseline results are \underline{underlined}, and the best-performing approach is highlighted in \textbf{bold}.} 
\label{tab:train_with_harmful_test_with_harmful}
\centering
\footnotesize 
\begin{tabular}{llrrrrrrrr}
\toprule
\multicolumn{1}{r}{$\downarrow$~\textbf{Train Dataset}} & \textbf{Test dataset$\rightarrow$} 
& \multicolumn{2}{c}{\textbf{Waseem}} & \multicolumn{2}{c}{\textbf{Davidson}} & \multicolumn{2}{c}{\textbf{Founta}} & \multicolumn{2}{c}{\textbf{HateXplain}} \\
\cmidrule(lr){3-4} \cmidrule(lr){5-6} \cmidrule(lr){7-8} \cmidrule(lr){9-10}
& & \textbf{R} & \textbf{F1} & \textbf{R} & \textbf{F1} & \textbf{R} & \textbf{F1} & \textbf{R} & \textbf{F1} \\

\midrule
\textbf{Waseem} & Original Dataset & - & - & \underline{0.743} & \underline{\textbf{0.836}} & \underline{0.597} & \underline{0.796} & \underline{0.547} & \underline{\textbf{0.76}} \\
       & + Drop Influential Samples & - & - & 0.781 & 0.759 & 0.69 & 0.805 & 0.768 & 0.744 \\
       & + GPT4o Influential Reannotation & - & - & 0.776 & 0.767 & 0.695 & 0.827 & 0.8 & 0.657 \\
       & + GPT4o + Llama3 Augmentation & - & - & \textbf{0.822} & 0.759 & \textbf{0.747} & \textbf{0.846} & \textbf{0.853} & 0.648 \\ 
       \addlinespace[0.1cm]
       \hdashline[1pt/4pt] 
        \addlinespace[0.1cm]
       & + GPT4o on Train and Test & - & - & 0.761 & 0.774 & 0.728 & 0.849 & 0.785 & 0.672 \\

\midrule
\textbf{Davidson} & Original Dataset & \underline{0.652} & \underline{0.7} & - & - & \underline{0.772} & \underline{0.81} & \underline{0.869} & \underline{\textbf{0.669}} \\
       & + Drop Influential Samples & \textbf{0.866} & 0.697 & - & - & \textbf{0.86} & 0.783 & \textbf{0.936} & 0.598 \\
       & + GPT4o Influential Reannotation & 0.702 & 0.724 & - & - & 0.798 & \textbf{0.832} & 0.922 & 0.627 \\
       & + GPT4o + Llama3 Augmentation & 0.764 & \textbf{0.729} & - & - & 0.826 & 0.824 & 0.914 & 0.616 \\
       \addlinespace[0.1cm]
       \hdashline[1pt/4pt] 
        \addlinespace[0.1cm]
       & + GPT4o on Train and Test & 0.71 & 0.752 & - & - & 0.826 & 0.849 & 0.909 & 0.637 \\
       
\midrule
\textbf{Founta} & Original Dataset  & \underline{0.596} & \underline{\textbf{0.742}} & \underline{0.874} & \underline{\textbf{0.852}} & - & - & \underline{0.784} & \underline{\textbf{0.724}} \\
       & + Drop Influential Samples & \textbf{0.68} & 0.737 & \textbf{0.895} & 0.805 & - & - & 0.813 & 0.717 \\
       & + GPT4o Influential Reannotation & 0.616 & 0.738 & 0.886 & 0.829 & - & - & \textbf{0.847} & 0.684 \\
       & + GPT4o + Llama3 Augmentation & 0.562 & 0.716 & 0.882 & 0.836 & - & - & 0.82 & 0.699 \\
        \addlinespace[0.1cm]
       \hdashline[1pt/4pt] 
        \addlinespace[0.1cm]
       & + GPT4o on Train and Test & 0.64 & 0.776 & 0.875 & 0.824 & - & - & 0.844 & 0.695 \\
\midrule
\textbf{HateXplain} & Original Dataset  & \underline{0.658} & \underline{0.704} & \underline{0.953} & \underline{0.726} & \underline{0.797} & \underline{\textbf{0.831}} & - & - \\
       & + Drop Influential Samples & 0.715 & \textbf{0.705} & 0.953 & \textbf{0.727} & 0.804 & \textbf{0.831} & - & - \\
       & + GPT4o Influential Reannotation & 0.762 & 0.695 & 0.957 & 0.711 & 0.814 & 0.822 & - & - \\
       & + GPT4o + Llama3 Augmentation & \textbf{0.777} & 0.665 & \textbf{0.961} & 0.699 & \textbf{0.827} & 0.811 & - & - \\
       \addlinespace[0.1cm]
       \hdashline[1pt/4pt] 
        \addlinespace[0.1cm]
       & + GPT4o on Train and Test & 0.762 & 0.731 & 0.962 & 0.728 & 0.851 & 0.838 & - & - \\
\bottomrule
\end{tabular}
\end{table*}

We defined three approaches for conducting the experiments and compare them against the baseline (training with the original dataset). The first approach involves training a generic model after removing all influential samples from the training dataset (see Table~\ref{tab:generic_datasets_looping_results}). The second approach involving reannotating the influential samples within each training dataset using GPT-4o, as detailed in Section~\ref{sec:gpt4_annotation_in_methodology}. The third approach involves augmenting the influential samples with paraphrased implicit versions generated by Llama-3 70B, as described in Section~\ref{sec:Llama_augmentation_in_methodology}. For each training dataset listed in Table~\ref{tab:train_with_harmful_test_with_harmful}, the last row (exploration) shows results where the influential samples were reannotated in both the training and testing datasets. This was done to explore the impact of testing the model on reannotated (cleaned) datasets.

\subsection{Generic Dataset as Training and Testing}
\begin{figure*}[ht]
\centering

\begin{subfigure}{0.24\textwidth}
\centering
\includegraphics[width=\linewidth]{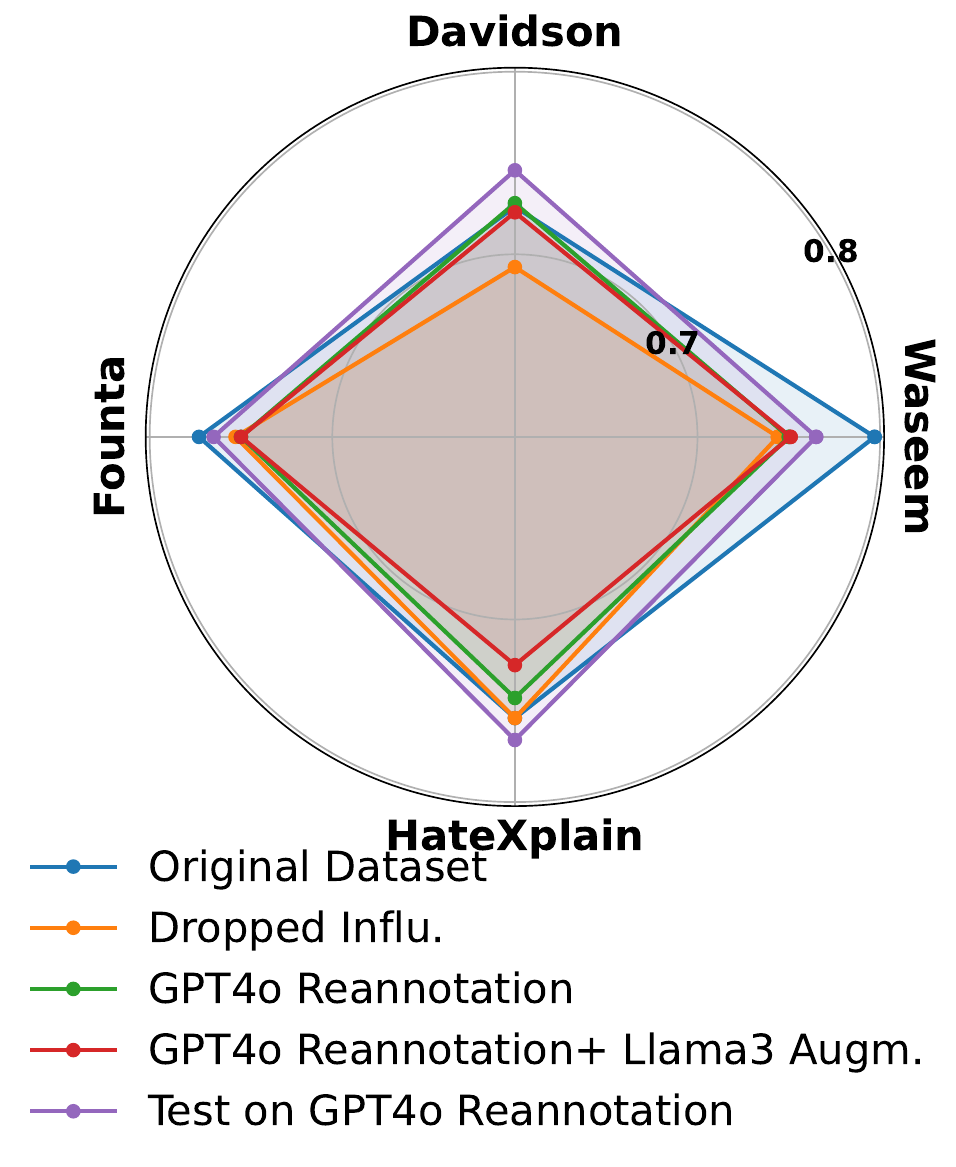}
\caption{Generic F1 Score}
\label{fig:Generic_F1}
\end{subfigure}\hfill
\begin{subfigure}{0.24\textwidth}
\centering
\includegraphics[width=\linewidth]{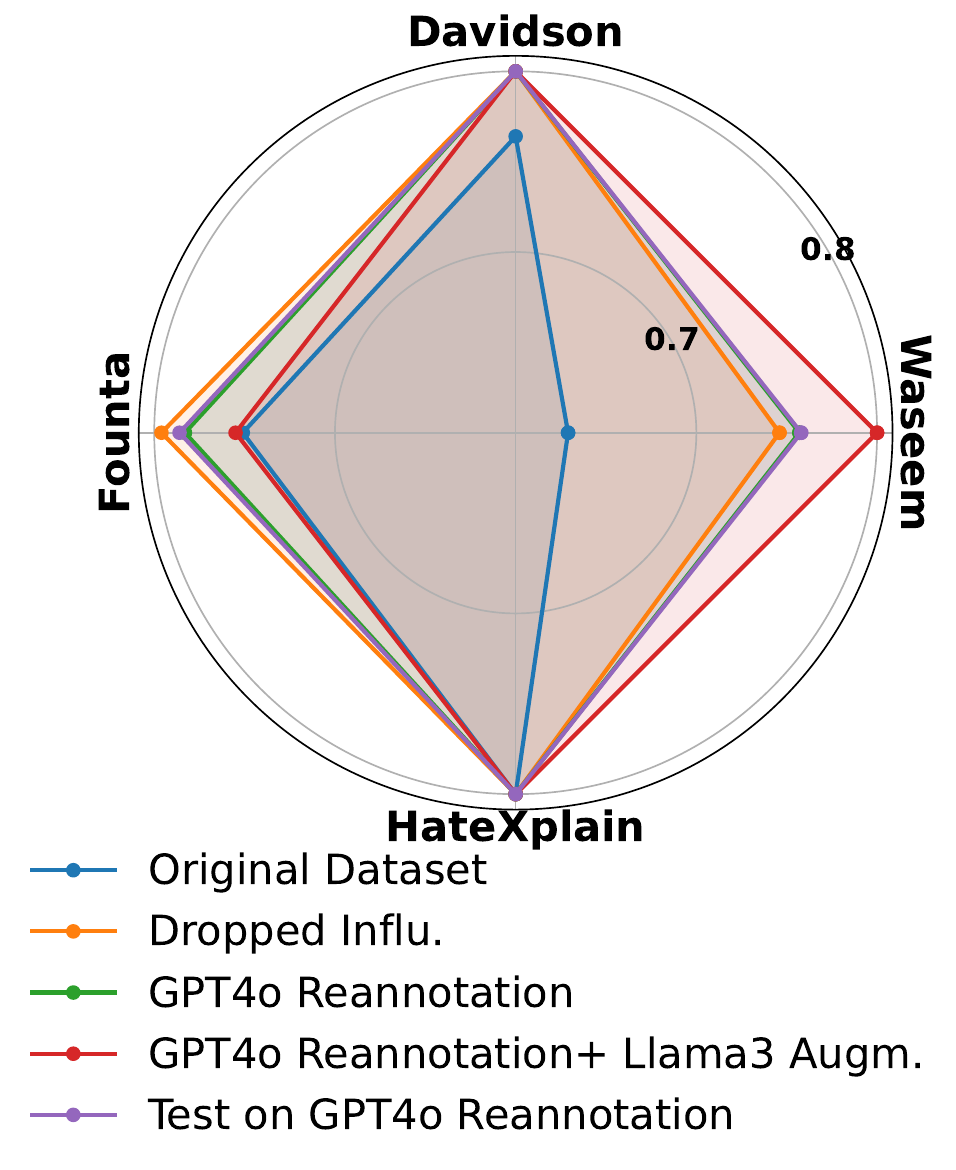}
\caption{Generic Recall Score}
\label{fig:Generic_Recall}
\end{subfigure}\hfill
\begin{subfigure}{0.24\textwidth}
\centering
\includegraphics[width=\linewidth]{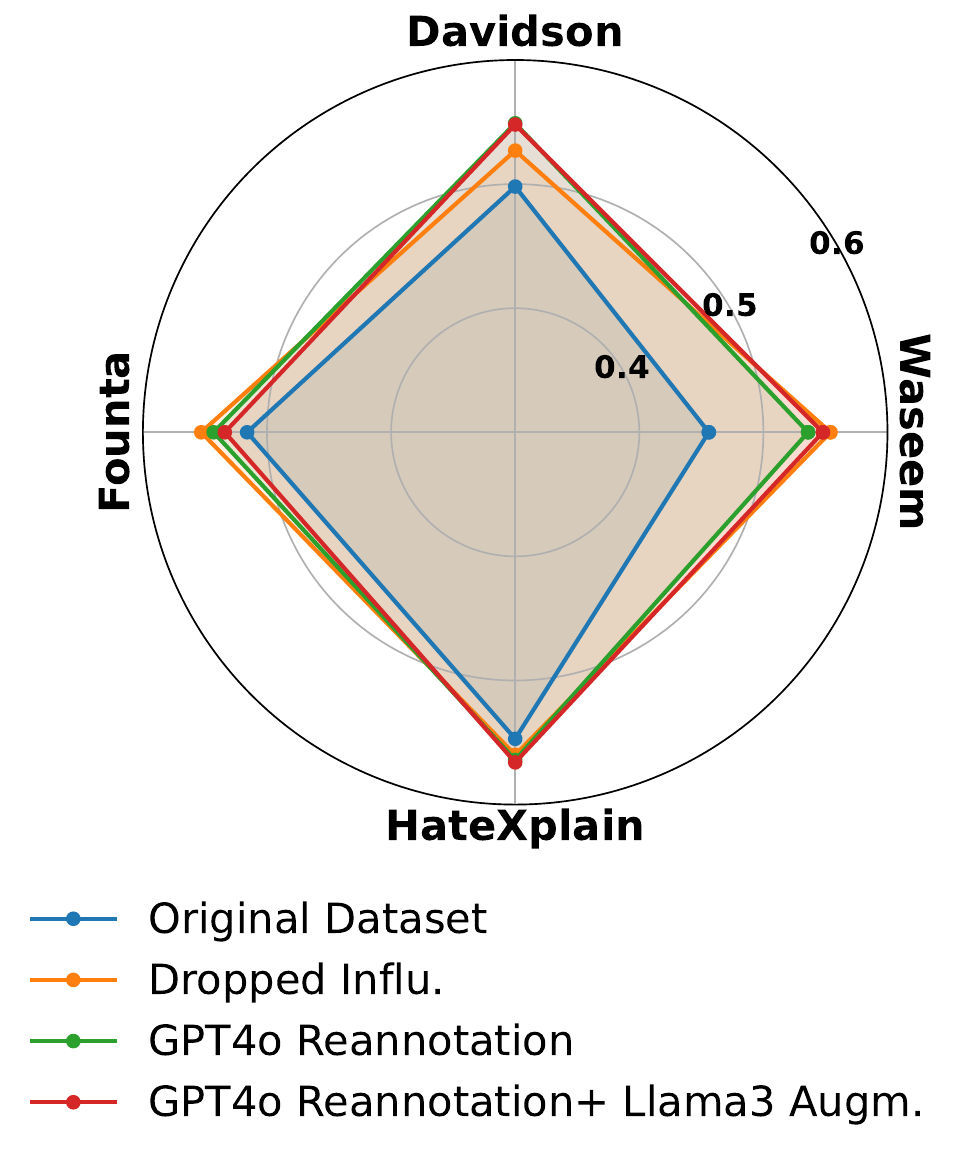}
\caption{Specialized F1 Score}
\label{fig:Specialized_F1}
\end{subfigure}\hfill
\begin{subfigure}{0.24\textwidth}
\centering
\includegraphics[width=\linewidth]{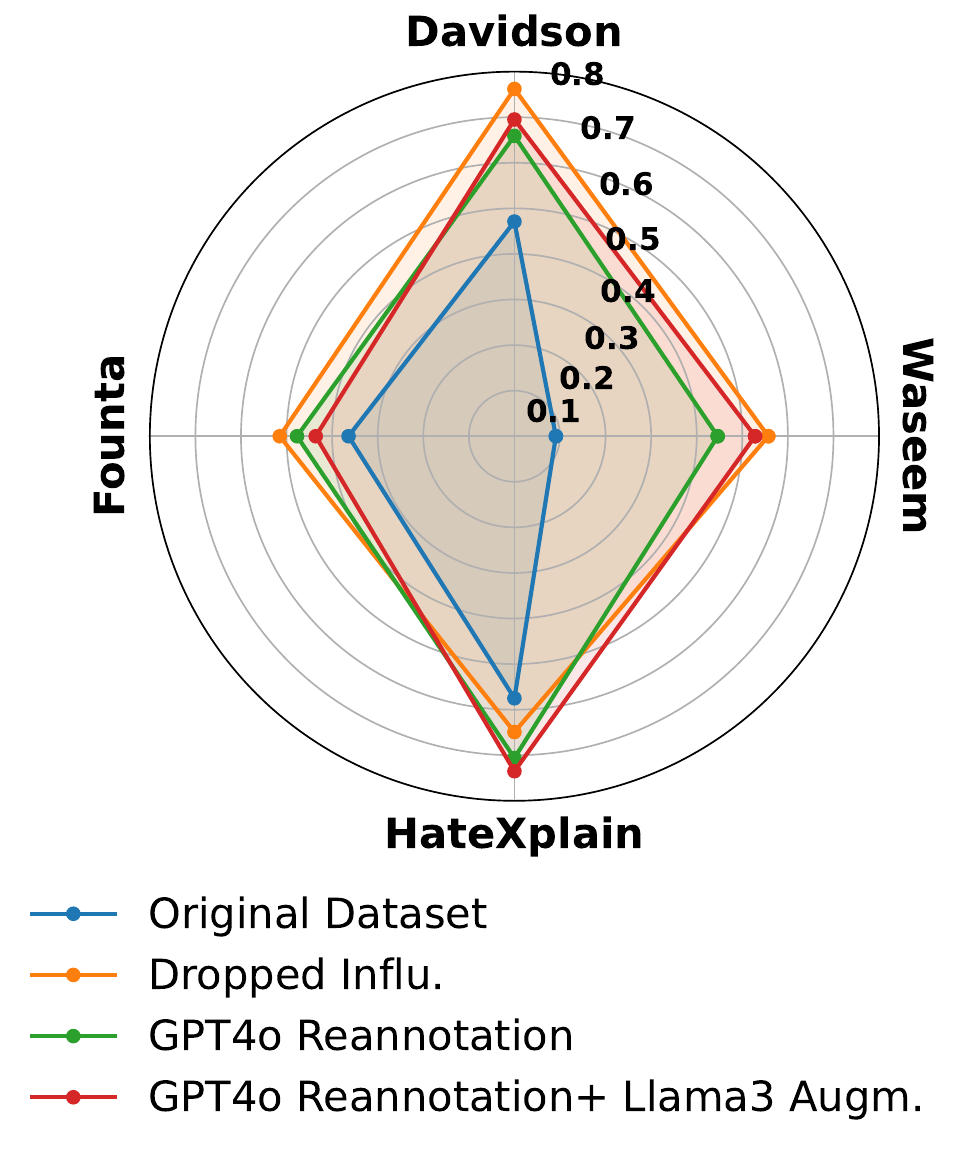}
\caption{Specialized Recall Score}
\label{fig:Specialized_Recall}
\end{subfigure}

\caption{Averaged Performance Metrics for training on generic datasets and evaluated over generic and specialized datasets (generic datasets in the figures represent the training dataset).}
\label{fig:All_Figures_Row}

\end{figure*}

In this experiment, the F1 score is more meaningful than Recall, although Recall is included for consistency across the tables. The F1 score is more relevant here because both the training and testing datasets classify any kind of harmful speech—whether explicit, implicit, or offensive—as part of the positive class. As shown in Table~\ref{tab:train_with_harmful_test_with_harmful} and Fig.~\ref{fig:Generic_F1}, the performance varies depending on the training dataset. The Waseem dataset performed best in the baseline. The Davidson dataset showed a preference for the second approach, achieving an average F1 score of 0.728 across the three test datasets. For the Founta dataset, the baseline outperformed other approaches by an average of 2 points in the F1 score. Finally, the HateXplain dataset performed slightly better with the first approach, though the results were nearly identical to the baseline.

\begin{table*}[h!]
\caption{Performance of generic datasets evaluated across specialized datasets. Baseline results are \underline{underlined}, and the best-performing approach is highlighted in \textbf{bold}.} \label{tab:train_with_harmful_test_with_implicit}
\centering
\footnotesize 
\begin{tabular}{llrrrrrr}
\toprule
\multicolumn{1}{r}{$\downarrow$~\textbf{Train Dataset}} & \textbf{Test dataset$\rightarrow$} 
& \multicolumn{2}{c}{\textbf{IHC}} & \multicolumn{2}{c}{\textbf{OLID\_IH}} & \multicolumn{2}{c}{\textbf{THOS\_IH}} \\
\cmidrule(lr){3-4} \cmidrule(lr){5-6} \cmidrule(lr){7-8}
& & \textbf{R} & \textbf{F1} & \textbf{R} & \textbf{F1} & \textbf{R} & \textbf{F1} \\

\midrule
\textbf{Waseem} & Original Dataset & \underline{0.064} & \underline{0.512} & \underline{0.14} & \underline{0.501} & \underline{0.069} & \underline{0.355} \\
       & + Drop Influential Samples & 0.64 & 0.588 & \textbf{0.564} & \textbf{0.59} & \textbf{0.467} & \textbf{0.484} \\
       & + GPT4o Influential Reannotation & 0.573 &\textbf{0.607} & 0.409 & 0.553 & 0.356 & 0.447 \\
       & + GPT4o + Llama3 Augmentation & \textbf{0.661} &\textbf{0.607} & 0.518 & 0.576 & 0.406 & 0.462 \\

\midrule
\textbf{Davidson} & Original Dataset 
                         & \underline{0.598} & \underline{0.563} & \underline{0.424} & \underline{0.509} & \underline{0.39} & \underline{0.422} \\
       & + Drop Influential Samples          &\textbf{0.851} & 0.562 & \textbf{0.789} & 0.55 & \textbf{0.647} & 0.469 \\
       & + GPT4o Influential Reannotation         & 0.8 & \textbf{0.607} & 0.649 & 0.566 & 0.527 & \textbf{0.475} \\
       & + GPT4o + Llama3 Augmentation  & 0.82 & 0.6 & 0.714 & \textbf{0.571} & 0.55 & 0.474 \\
       
\midrule
\textbf{Founta} & Original Dataset
                         & \underline{0.459} & \underline{0.586} & \underline{0.378} & \underline{0.546} & \underline{0.256} & \underline{0.415} \\
       & + Drop Influential Samples          & 0.535 & 0.582 & \textbf{0.58} & \textbf{0.594} & \textbf{0.43} & \textbf{0.482} \\
       & + GPT4o Influential Reannotation         & \textbf{0.612} & \textbf{0.611} & 0.476 & 0.57 & 0.344 & 0.448 \\
       & + GPT4o + Llama3 Augmentation  & 0.544 & 0.6 & 0.459 & 0.569 & 0.306 & 0.432 \\
\midrule
\textbf{HateXplain} & Original Dataset 
                         & \underline{0.701} & \underline{\textbf{0.62}} & \underline{0.58} & \underline{0.56} & \underline{0.443} & \underline{0.462} \\
       & + Drop Influential Samples          & 0.721 & 0.605 & 0.68 & 0.571 & 0.545 & 0.504 \\
       & + GPT4o Influential Reannotation         & 0.82 & 0.609 & 0.712 & 0.571 & 0.587 & 0.511 \\
       & + GPT4o + Llama3 Augmentation  & \textbf{0.824} & 0.6 & \textbf{0.768} & \textbf{0.577} & \textbf{0.614} & \textbf{0.521} \\
\bottomrule
\end{tabular}
\end{table*}

\begin{table}[h!]
\caption{Performance of specialized datasets evaluated across each other (for comparison with the performance of the generic datasets in Table~\ref{tab:train_with_harmful_test_with_implicit}).} 

\label{tab:train_with_implicit_test_with_implicit}
\centering
\footnotesize 
\resizebox{\columnwidth}{!}{%
\begin{tabular}{llrrrrrr}
\toprule
\multirow{2}{*}{\makecell[c]{\textbf{Test Dataset}~$\rightarrow$ \\ \textbf{Train Dataset}~$\downarrow$}} &

\multicolumn{2}{c}{\textbf{IHC}} & \multicolumn{2}{c}{\textbf{OLID\_IH}} & \multicolumn{2}{c}{\textbf{THOS\_IH}} \\
\cmidrule(lr){2-3} \cmidrule(lr){4-5} \cmidrule(lr){6-7}
 & \textbf{R} & \textbf{F1} & \textbf{R} & \textbf{F1} & \textbf{R} & \textbf{F1} \\

\midrule
\textbf{IHC}   & - & - & 0.659 & 0.557 & 0.712 & 0.529 \\

\midrule
\textbf{OLID\_IH}   & 0.136 & 0.529 & - & -  & 0.077 & 0.516 \\
       
\midrule
\textbf{THOS\_IH}  & 0.047 & 0.51 & 0.016 & 0.502 & - & - \\

\bottomrule
\end{tabular}
}
\end{table}

\subsection{Generic Dataset as Training and Specialized as Testing}
In this experiment, the Recall metric is more meaningful than the F1 score because the generic-trained model is expected to produce a high number of False Positives (FP) when tested on implicit hate datasets, where explicit harmful speech is labeled as part of the negative class. As shown in Table~\ref{tab:train_with_harmful_test_with_implicit} and Fig.~\ref{fig:Specialized_Recall}, the Waseem, Davidson, and Founta datasets favored the first approach, achieving the best performance on implicit hate testing. In contrast, HateXplain performed best with the third approach.

Regarding model generalizability, Fig.~\ref{fig:Specialized_F1} and~Fig.~\ref{fig:Specialized_Recall} illustrate that all proposed approaches substantially improved performance compared to training on the original datasets. This demonstrates the effectiveness of the proposed methods in generalizing generic datasets for implicit hate detection in specialized datasets. Table~\ref{tab:train_with_implicit_test_with_implicit} further highlights the performance of specialized datasets in cross-dataset testing scenarios. Despite the limited availability of implicit hate datasets, we evaluated their effectiveness when the model was trained and tested on the same type of dataset, comparing the results with our generalized approaches. The results indicate that THOS\_IH and OLID\_IH struggled to perform well in cross-dataset testing, whereas the IHC dataset showed stronger performance. However, when examining the average Recall across tests with OLID\_IH and THOS\_IH, Davidson in the first approach and HateXplain in the third approach outperformed the specialized IHC dataset. Additionally, as shown in Table~\ref{tab:train_with_harmful_test_with_implicit}, the remaining results were comparable to IHC's performance in Table~\ref{tab:train_with_implicit_test_with_implicit}. In terms of the F1 score, HateXplain under the third approach also outperformed IHC.

\subsection{Reannotated Dataset as Training and Testing}
As shown in the last row of each training dataset section in Table~\ref{tab:train_with_harmful_test_with_harmful}, 7 out of 12 experiments achieved higher F1 scores compared to all other approaches, including the baseline. The results show that reannotating Davidson and HateXplain on testing led to a decrease in performance. In contrast, reannotating Waseem and Founta as testing datasets resulted in improved performance in all experiments, outperforming other approaches, including the baseline. This discrepancy raises important questions. Given that Waseem and Founta have a high proportion of influential samples, as shown in Table~\ref{tab:generic_datasets_looping_results}, it is possible that these datasets initially contained significant noise, which was corrected through relabeling. Conversely, this result may suggest that HateXplain and Davidson require additional training loops, as illustrated in our pipeline (Fig.~\ref{fig:pipeline_fig}), to identify and correct more influential samples for further improvements.

From a different perspective, when comparing the exploration approach, in which GPT-4o reannotation was applied to both training and testing data, with the second approach, where GPT-4o reannotation was applied only to the training data, the exploration approach yielded better performance in all experiments. The only exception occured when the model was trained with Founta and tested with Davidson, where the second approach achieved an F1 score of 0.829, slightly higher than the 0.824 score for the exploration approach.

\section{Conclusion}

In this paper, we proposed an approach to generalize generic datasets toward a subtle and previously unannotated class by leveraging a trusted samples dataset generated with GPT-4o under the guidance and curation of two experts. The approach involves identifying influential samples in the training data and applying various configurations, such as removing influential samples, GPT-4o reannotation, and Llama-3 augmentation. To evaluate the performance of our proposed approach, we conducted experiments using seven datasets of  harmful speech. Among these, four are generic datasets focusing on hate speech and offensive language, while three are specialized datasets on implicit hate speech. Our results demonstrate the effectiveness of the proposed approach to generalize the explicit hate datasets to classify implicit hate samples, achieving a 12.9\% improvement in F1 score on specialized datasets while maintaining comparable performance on the generic datasets.
\section*{Limitations}
In our second and third approaches, where we engaged the LLM to determine whether a given influential sample is mislabeled, the solution remains suboptimal. As observed in previous research \cite{almohaimeed2024transfer}, \cite{gpt4-donmez2024please}, LLMs have not yet reached the level of human experts in accurately identifying harmful content, particularly in its implicit form. This limitation may lead to the mislabeling of critical data, posing a challenge to the reliability of the model.

Additionally, in our methodology pipeline, the selection of the best-performing version of the training dataset after removing a set of influential samples is not automated. Instead, the choice was made based on our manual observation of the optimal loop results. Developing an approach to systematically determine whether a given version yields the best results would enable a fully automated pipeline. Such an advancement could be beneficial for future research and facilitate the development of tools for both academic and production settings.

\newpage

\bibliography{references}
\bibliographystyle{acl_natbib}

\newpage
\appendix
\section*{Appendix A: Dataset Licenses}
\label{appendix:licenses}

Table~\ref{tab:datasets_licenses} lists the datasets used in this paper along with their associated usage licenses or permission conditions.

\begin{table}[h!]
\caption{Licenses and usage conditions of the datasets used in this paper. \textbf{P} indicates datasets with no explicit license but made publicly available by the authors with a request for citation. \textbf{MIT} refers to the MIT License. \textbf{CC} denotes the Creative Commons Attribution 4.0 (CC BY 4.0) license, and \textbf{CCNC} refers to the Creative Commons Attribution-NonCommercial-ShareAlike 4.0 International (CC BY-NC-SA 4.0) license.
}\label{tab:datasets_licenses}
\centering
\resizebox{\columnwidth}{!}{%
\begin{tabular}{lll}
\toprule
\textbf{Dataset} &\textbf{License} \\
\midrule
Waseem~\cite{waseem-hovy:2016:N16-2}& P\\
Davidson~\cite{davidson2017automated} & MIT \\
Founta~\cite{founta2018large} & CC\\
HateXplain~\cite{hatexplain-mathew2021} & MIT\\
IHC~\cite{IHC-elsherief-etal-2021-latent} & MIT\\
OLID\_IH~\cite{OLID_IH-caselli-etal-2020-feel}& CCNC\\
THOS\_IH~\cite{THOS-almohaimeed} & CCNC\\

\bottomrule
\end{tabular}
}
\end{table}

\end{document}